\pdfoutput=1
\documentclass[11pt]{article} 
\usepackage{naacl}

\usepackage{amsmath,amsfonts,bm}

\def\eqref#1{equation~\ref{#1}}

\def\1{\bm{1}}

\DeclareMathAlphabet{\mathsfit}{\encodingdefault}{\sfdefault}{m}{sl}
\SetMathAlphabet{\mathsfit}{bold}{\encodingdefault}{\sfdefault}{bx}{n}

\usepackage[utf8]{inputenc}
\usepackage{subfigure}
\usepackage{epsfig}
\usepackage{color}
\usepackage{microtype}
\usepackage{multirow}
\usepackage{amsfonts}
\usepackage{amssymb}
\usepackage{amsmath}
\usepackage{booktabs}
\usepackage{xcolor}
\usepackage{bm}
\usepackage{numprint}
\usepackage{forest}
\usepackage{tikz-qtree}
\usepackage{url}
\usepackage{times}
\usepackage{latexsym}
\usepackage{bbding}
\usepackage{tikz-qtree}
\usepackage{appendix}
\usepackage{wrapfig}
\usepackage{stfloats}

\newcommand{\lmethodname}{LC-PCFG}

\definecolor{bleudefrance}{rgb}{0.0, 0.36, 1.0}

\usepackage{times}
\usepackage{latexsym}

\usepackage[T1]{fontenc}

\usepackage[utf8]{inputenc}

\usepackage{microtype}

\usepackage{inconsolata}

\usepackage{amssymb}
\usepackage{pifont}
\newcommand{\xmark}{\ding{55}}%

\title{Re-evaluating the Need for Multimodal Signals \\ in Unsupervised Grammar Induction}

\author{
	Boyi Li\textsuperscript{*, 1} \qquad
	Rodolfo Corona\textsuperscript{*, 1} \qquad
	Karttikeya Mangalam\textsuperscript{*, 1}  \qquad  
	\textbf{Catherine Chen}\textsuperscript{*, 1}  \vspace{0.5em} \\
	\textbf{Daniel Flaherty}\textsuperscript{1} \qquad \quad  
        \textbf{Serge Belongie}\textsuperscript{2} \quad 
        \textbf{Kilian Q. Weinberger}\textsuperscript{3}  \quad \vspace{0.5em} \\ 
        \textbf{Jitendra Malik}\textsuperscript{1} \quad 
        \textbf{Trevor Darrell}\textsuperscript{1} \quad 
        \textbf{Dan Klein}\textsuperscript{1} \quad 
 \vspace{.8em} \\
	\textsuperscript{1}UC Berkeley \qquad \qquad \textsuperscript{2}University of Copenhagen  \qquad \qquad 
    \textsuperscript{3}Cornell University
	 \\
} 
\definecolor{demphcolor1}{gray}{.5}

\usepackage{siunitx}

\newcommand{\stdev}[1]{\textcolor{demphcolor1}{\tiny{\pm #1}}}

\begin{document}

\maketitle
\renewcommand*{\thefootnote}{\fnsymbol{footnote}}
    \setcounter{footnote}{1}
    \renewcommand*{\thefootnote}{\arabic{footnote}}
    \setcounter{footnote}{0}
\begin{abstract}
Are multimodal inputs necessary for grammar induction? Recent work has shown that multimodal training inputs can improve grammar induction. However, these improvements are based on comparisons to weak text-only baselines that were trained on relatively little textual data. To determine whether multimodal inputs are needed in regimes with large amounts of textual training data, we design a stronger text-only baseline, which we refer to as \lmethodname.  \lmethodname{} is a C-PFCG that incorporates embeddings from text-only large language models (LLMs). We use a fixed grammar family to directly compare \lmethodname{} to various multimodal grammar induction methods. We compare performance on four benchmark datasets. \lmethodname~provides an up to 17\% relative improvement in Corpus-F1 compared to state-of-the-art multimodal grammar induction methods. \lmethodname{} is also more computationally efficient, providing an up to $85\%$ reduction in parameter count and $8.8\times$ reduction in training time compared to multimodal approaches. These results suggest that multimodal inputs may not be necessary for grammar induction, and emphasize the importance of strong vision-free baselines for evaluating the benefit of multimodal approaches.
\end{abstract}

\section{Introduction}
\label{intro}

Prior studies have shown that multimodal inputs can facilitate grammar induction. These studies paired text with inputs from images and videos, and found that models trained with paired multimodal inputs outperform text-only models on grammar induction \cite{shi2019visually,zhao-titov-2020-visually,zhang2021video,zhang2022training}. These results suggest that multimodal inputs improve grammar induction by grounding textual inputs to the visual world. Indeed, a long line of work in human language learning suggests that paired multimodal inputs are crucial for language acquisition in humans \cite{gleitman1990structural,pinker1984language}. While multimodal inputs can undoubtedly help with grammar induction, especially in regimes with low textual data, are multimodal inputs \textit{necessary} to learn a grammar? To investigate this question, we test whether the benefits of multimodal inputs for grammar induction can be achieved by more textual data.

Prior studies of multimodal grammar induction compared multimodal methods to weak text-only baselines which were trained with relatively little data \cite{shi2019visually,zhao-titov-2020-visually,zhang2021video,zhang2022training}. However, recent grammar induction approaches that incorporate representations from large language models (LLMs) produced large improvements in text-only grammar induction performance \citep[e.g.,][]{cao2020unsupervised,drozdov2019unsupervised,li2023contextual}. The performance of these LLM-based grammar induction methods suggest that exposure to larger quantities of textual training data can substantially improve grammar induction. However, prior studies used different settings to evaluate multimodal and text-only methods for grammar induction. Thus, it is unclear whether the performance of LLM-based grammar induction approaches can match the performance of multimodal approaches.

\begin{figure} 
    \centering
     \includegraphics[width=\linewidth]{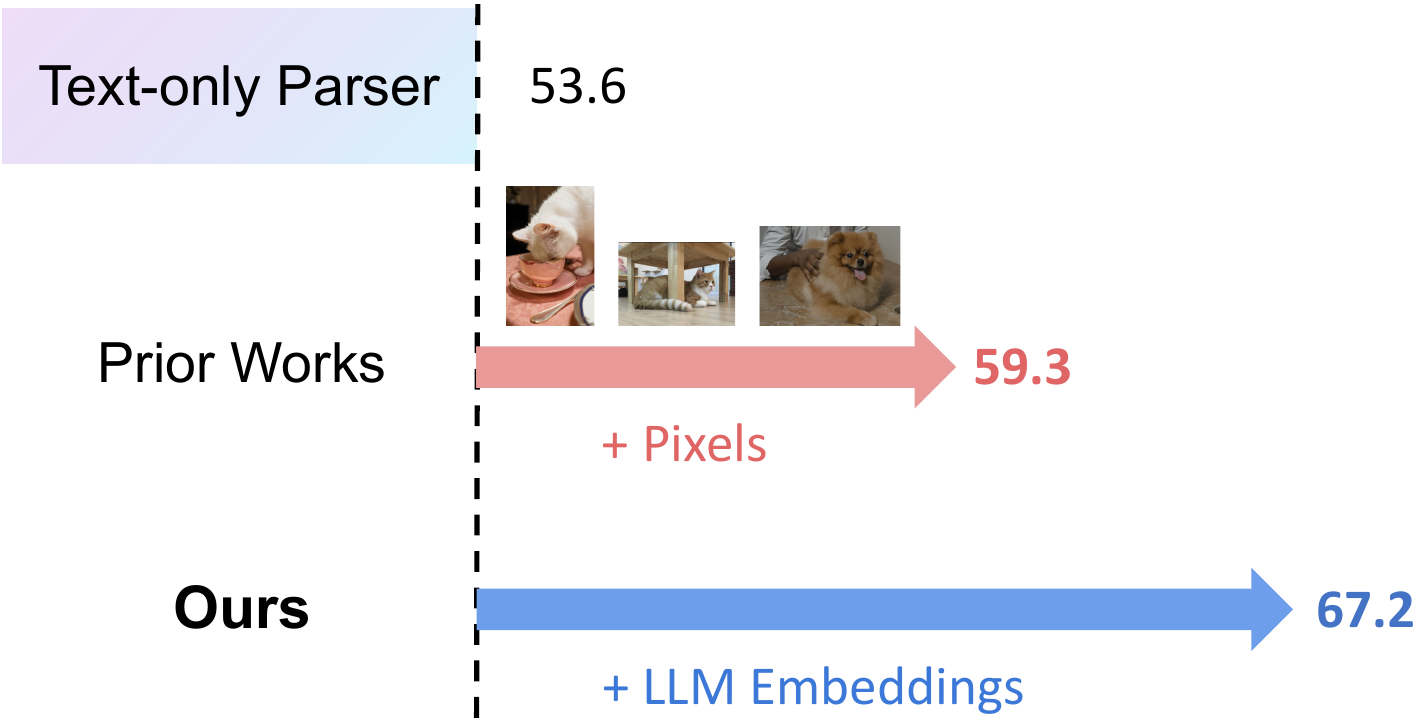}
    \caption{Comparison with prior multimodal methods on image-assisted grammar induction. Prior works showed that paired images can improve grammar induction ($53.6 \rightarrow 59.3$ Corpus-level F1). We show that a strong text-only baselined line that incorporates embeddings from large language models (LLM) can match (and surpass) multimodal methods, suggesting that multimodal inputs may not be necessary for grammar induction ($53.6 \rightarrow 67.2$). }
    \label{fig:teaser}
    \vspace{-0.2in}
\end{figure}

Here we compare multimodal methods for grammar induction to a strong text-only baseline. Our text-only baseline, which we refer to as \lmethodname, is a C-PCFG that incorporates embeddings from text-only LLMs. We use and use a fixed grammar family (C-PCFGs) to directly compare \lmethodname{} to multimodal methods, and perform comparisons with four multimodal grammar induction datasets. We find that compared to previous state-of-the-art multimodal methods, \lmethodname{} achieves up to 17\% relative improvement in Corpus-F1 score while requiring $8.8\times$ less time to train. Moreover, the benefits of incorporating LLM embeddings does not straightforwardly stack with the benefits of multimodal training inputs: adding multimodal training inputs to \lmethodname{} does not improve performance on grammar induction, suggesting that the benefits of multimodal inputs may be subsumed by training on large quantities of text. While multimodal training inputs may be useful in some settings, our results suggest that grammar induction may not require multimodal inputs. To facilitate further research we release our code at \href{https://github.com/Boyiliee/Vision-free-Multimodal-Grammar-Induction}{https://github.com/Boyiliee/Vision-free-Multimodal-Grammar-Induction}.

\section{Related Work}
\label{related_work}

\paragraph{Grammar induction} Grammar induction, the task of inducing syntactic structure without explicit supervision, has been extensively studied over the past few decades \cite[e.g.,][]{lari1990estimation,carroll1992two, Clark2001UnsupervisedIO,klein2002generative,smith2005guiding}.

Many methods for grammar induction train on data from text alone \cite[e.g.,][]{lari1990estimation,carroll1992two,klein2002generative,Shen2018NeuralLM,shen2019ordered}. However, based on the intuition that multimodal inputs capture information that is missing in text, recent studies have devised methods for grammar induction that incorporate information from images and videos~\citep{shi2019visually, zhao-titov-2020-visually,zhang2021video,zhang2022training}. These multimodal methods have been shown to outperform some text-only methods \citep{zhao-titov-2020-visually,shi2019visually,zhang2022training,zhang2021video}.

\paragraph{LLM features for grammar induction.} Recent advances in LLMs have enabled vast improvements on a wide range of downstream tasks, including both supervised syntactic parsing and grammar induction \cite[e.g.,][]{Devlin2019BERTPO,radford2019language,Kitaev2018MultilingualCP,cao2020unsupervised,drozdov2019unsupervised,li2023contextual}. However, prior work that evaluated the benefit of multimodal inputs for grammar induction used text-only baselines that incorporated much weaker word representations, such as random word embeddings or lexical word embeddings such as fastText. Thus, it is unclear whether stronger text-only methods for grammar induction can match the performance of multimodal approaches.

\section{\lmethodname{}: Grammar Induction with Large Language Models}\label{sec:method}

\begin{figure*}
    \centering
     \includegraphics[width=0.8\linewidth]{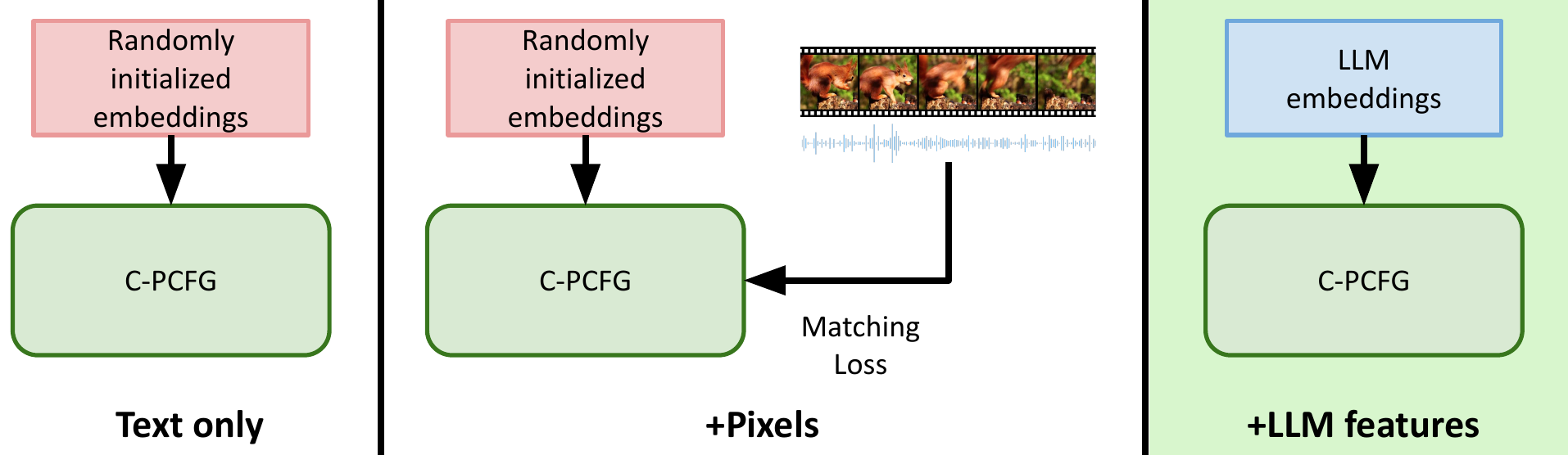}
    \caption{\textbf{Experimental Settings.} We explore using large language model features for unsupervised grammar induction. We use three experimental settings. (1) the standard setting in which word representations are learned from scratch (\textbf{Text Only}), (2) prior methods that incorporate a multimodal regularization loss (\textbf{+Pixels}), and (3) our method, which uses pre-trained text-only LLM features (\textbf{+LLM features}). We show that LLM features can obtain state-of-the-art performance, without requiring multimodal regularization.}
    \label{fig:experimental_conditions}
\end{figure*}

The goal of grammar induction is to learn syntactic structure without explicit supervision. Methods for grammar induction assume a grammar formalism and then optimize grammar parameters to fit the data. We use Compound Probabilistic Context-Free Grammars (C-PCFGs)~\citep{kim-etal-2019-compound} as a grammar formalism. We construct a C-PCFG that incorporates LLM representations. We refer to this method as \lmethodname{}. We compare \lmethodname{} to prior methods that incorporate multimodal data~\citep{zhao-titov-2020-visually,zhang2021video,zhang2022training}. Figure \ref{fig:experimental_conditions} provides an overview of our experiments.

\begin{figure}
    \centering
     \includegraphics[width=0.7\linewidth]{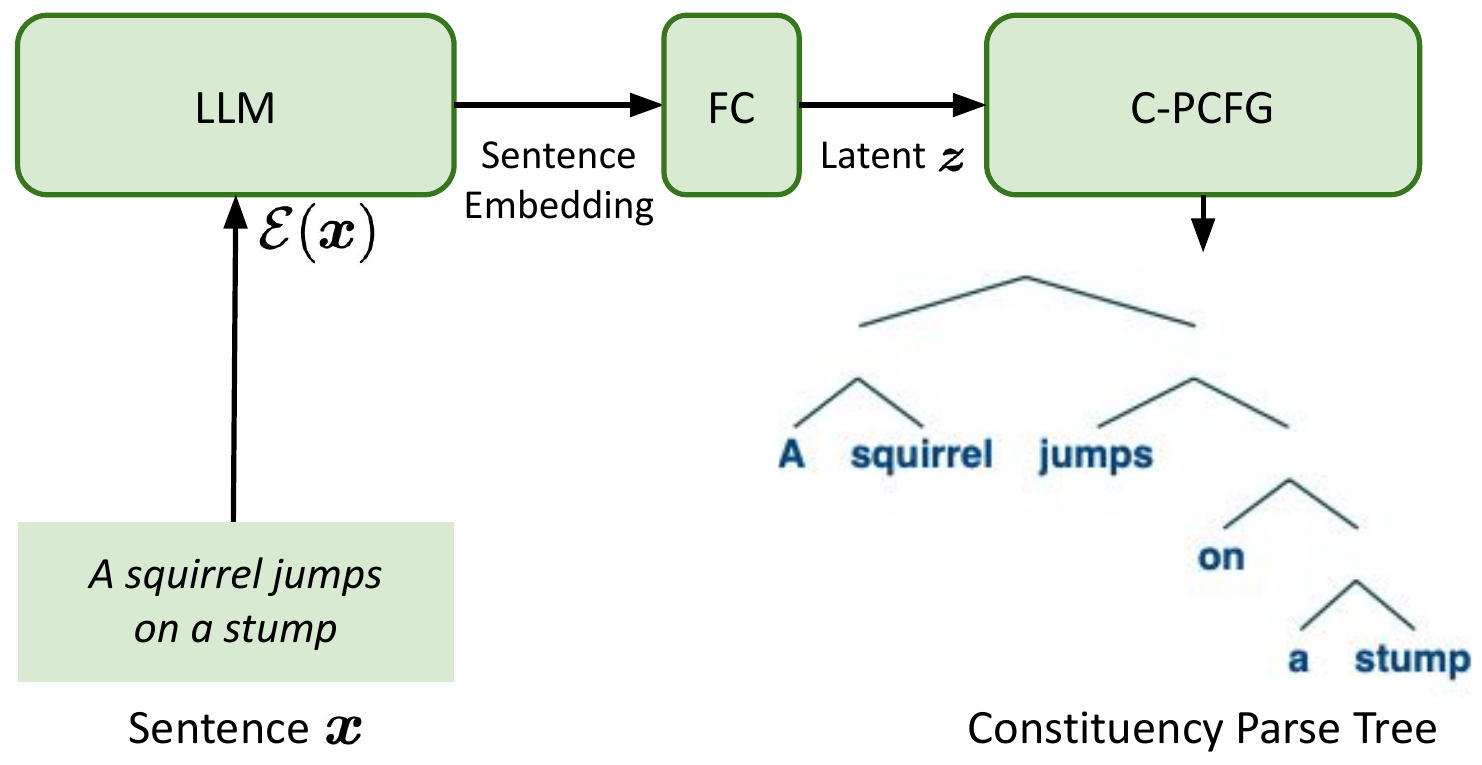}
    \caption{\textbf{\lmethodname{} workflow.} A sentence $\bm{x}$ is fed to an LLM to obtain a sentence embedding $\mathcal{E}(\bm{x})$. $\mathcal{E}(\bm{x})$ is passed through a fully-connected layer (FC), producing the latent $\bm{z}$. $\bm{z}$ is fed to the C-PCFG to obtain a constituency parse tree. Note that unlike prior work, our approach does not require multimodal data. }
    \label{fig:method}
    \vspace{-0.1in}
\end{figure}

\paragraph{Background.} C-PCFGs extend the Probabilistic Context Free Grammar (PCFG) formalism, and are defined by a 5-tuple $\mathcal{G}=(S,\mathcal{N},\mathcal{P},\Sigma,\mathcal{R})$, consisting of a start symbol $S$, a set of non-terminals $\mathcal{N}$, a set of pre-terminals $\mathcal{P}$, a set of terminals $\Sigma$, and a set of derivation rules $\mathcal{R}$:
\begin{align*}
S\to A && A\in \mathcal{N} \\
A\to BC && A\in \mathcal{N}, B,C\in \mathcal{N}\cup \mathcal{P} \\
T\to w && T\in \mathcal{P}, w\in \Sigma 
\end{align*}
PCFGs define a probability distribution over transformation rules $\bm{\pi}=\{\pi_r\}_{r\in\mathcal{R}}$. Then the inside algorithm~\citep{baker1979trainable} can be used to efficiently perform inference over this probability distribution. 
In neural PCFGs, this distribution may be formulated as follows: 
\begin{align*}
    \pi _{S\to A}=\frac{\text{exp}(\bm{u}_A^{\top} f_1(\bm{w}_S))}{\Sigma _{A'\in\mathcal{N}}\text{exp}(\bm{u}_{A'}^\top f_1(\bm{w}_S))} \\
    \pi _{A\to BC}=\frac{\text{exp}(\bm{u}_{BC}^{\top} \bm{w}_A)}{\Sigma _{B'C'\in\mathcal{M}}\text{exp}(\bm{u}_{B'C'}^\top \bm{w}_A)} \\
    \pi _{T\to w}=\frac{\text{exp}(\bm{u}_w^{\top} f_2(\bm{w}_T))}{\Sigma _{w'\in\Sigma}\text{exp}(\bm{u}_{w'}^\top f_2(\bm{w}_T))} \\
\end{align*}
where $\bm{u}$ are transformation vectors for each production rule, $\bm{w}$ are learnable parameter vectors for each symbol, and $f_1$ and $f_2$ are neural networks. 
The neural PCFG formulation preserves the benefits of fast inference while additionally incorporating distributional representations from neural networks. 

Because PCFGs contain a strong context-free assumption, PCFGs cannot leverage global information that is useful for computing production probabilities during inference. C-PCFGs~\citep{kim-etal-2019-compound} extend PCFGs to incorporate global information. C-PCFGs formulate rule probabilities as a compound probability distribution~\citep{robbins1956empirical}:
\begin{align*}
    \bm{z}\sim p_{\gamma}(\bm{z}) && \pi _{\bm{z}} = f_{\lambda}(\bm{z}, \bm{E}_{\mathcal{G}})
\end{align*}
Where $\bm{z}$ is a latent variable generated by a prior distribution (generally assumed to be spherical Gaussian) and $\bm{E}_{\mathcal{G}}=\{\bm{w}_N|N\in \{S\}\cup \mathcal{N} \cup \mathcal{P}\}$ denotes the set of symbol embeddings. 
Rule probabilities $\bm{\pi}_{\bm{z}}$ are additionally conditioned on this latent variable: 
\begin{align*}
    \pi _{\bm{z}, S\to A}\propto \text{exp}(\bm{u}_{A}^\top f_1 ([\bm{w}_S;\bm{z}])), \\
    \pi _{\bm{z}, A\to BC}\propto \text{exp}(\bm{u}_{BC}^\top [\bm{w}_A;\bm{z}]), \\
    \pi _{\bm{z}, T\to w}\propto \text{exp}(\bm{u}_{w}^\top f_2 ([\bm{w}_T;\bm{z}]))
\end{align*}
Importantly, the latent variable $\bm{z}$ allows global information to be shared across production decisions during, while maintaining the context-free assumption needed for efficient inference when $\bm{z}$ is fixed. 

Becase the introduction of $\bm{z}$ makes inference intractable, variational methods are used to optimize C-PCFGs~\citep{kingma2013auto}
At inference time, given a sentence $\bm{x}$, the variational inference network $q_\phi$ is used to produce the latent $\bm{z} = \bm{\mu}_\phi (g(\mathcal{E}(\bm{x})))$.
Here, $g$ is a sentence encoder used to generate a vector representation given token embeddings $\mathcal{E}(\bm{x})$.
For a more thorough treatment of C-PCFGs, please see \citet{kim-etal-2019-compound}.

\paragraph{LLM-based C-PCFG for grammar induction.}
We design \lmethodname{}, a simple but strong text-only baseline which incorporates pre-trained LLM representations into the C-PCFG inference network. 
Specifically, we formulate the inference network as: 
\begin{align}
    \mathcal{E}(\bm{x}) &= \text{LLM}(\bm{x}) \\
    g(\bm{x}) &= \text{FC}(m(\mathcal{E}(\bm{x})))
\end{align}
where $m$ represents a mean-pool operation. 
Here, an LLM is used to obtain text embeddings for each sentence $\bm{x}$, which are then fed to a fully connected (FC) layer as the C-PCFG inference network. Figure~\ref{fig:method} provides an example of this method. A sentence $\bm{x}$ (``\textit{A squirrel jumps on a stump}") is fed into an LLM to obtain an embedding of the sentence. Then the sentence embedding is passed into a fully-connected layer to obtain the latent variable $\bm{z}$. Finally, we feed $\bm{z}$ into the C-PCFG to obtain a constituency parse tree. 
Note that compared to prior multimodal CPFGs which used multimodal inputs for regularization, our approach does not use any multimodal data.

\begin{table*}[htb!]\small
	\centering
	\caption{
    \label{tab:eval_mscoco}
    \textbf{Grammar induction with image and text}. Corpus-level F1 (C-F1) and sentence-level F1 (S-F1) scores on the MSCOCO 2014 caption dataset.
   We compare LC-PCFG against simple rule-based baselines (top, from~\cite{zhao-titov-2020-visually}), prior state-of-the-art methods that employ image data (middle), and methods, including ours, that use purely textual data (bottom). 
   \textit{RGB} indicates whether each method uses multimodal inputs.
    LC-PCFG outperforms all prior multimodal methods. 
}

\begin{tabular}{ll|ccc|ll}
\toprule
\phantom{text} & Method & RGB & LLM & Params (M) & C-F1 & S-F1 \\
\midrule
\multicolumn{4}{l}{\textit{Rule-based baselines}} & \\ 
\rule{0pt}{\normalbaselineskip} 
& Left Branching & No & No & ~~~- & $15.1$ & $15.7$ \\
& Right Branching & No & No & ~~~- & $51.0$ & $51.8$ \\
& Random Trees & No & No & ~~~- & $24.2_{\stdev{0.3}}$ & $24.6_{\stdev{0.2}}$ \\
\midrule
\multicolumn{4}{l}{\textit{Methods using extra-linguistic inputs}} & \\ 
\rule{0pt}{\normalbaselineskip} 
& VG-NSL~\citep{shi2019visually} & Yes & No & ~~~- & $50.4_{\stdev{0.3}}$ & ~~~- \\
& VC-PCFG~\citep{zhao-titov-2020-visually} & Yes & No & $41.5$ & $59.3_{\stdev{8.2}}$ & $59.4_{\stdev{8.3}}$\\
& VC-PCFG++ & Yes & No & $41.5$ & $64.2_{\stdev{7.0}}$ & $64.6_{\stdev{7.2}}$\\
\midrule
\multicolumn{4}{l}{\textit{Methods using only textual inputs}} & \\ 
\rule{0pt}{\normalbaselineskip} 
& C-PCFG~\citep{kim-etal-2019-compound} & No & No & $15.3$ & $53.6_{\stdev{4.7}}$ & $53.7_{\stdev{4.6}}$\\
& \textbf{LC-PCFG} (Ours) & No & Yes & $\textbf{6.2}$ & $\textbf{67.2}_{\stdev{1.1}}$ & $\textbf{67.8}_{\stdev{1.2}}$ \\
\bottomrule
\end{tabular}

 \vspace{-0.2in}
\end{table*}

\section{Experiments}\label{sec:experiments}
\subsection{Image-assisted Parsing}
\label{subsec:image-assisted-parsing}
We compare \lmethodname{} against VG-NSL~\citep{shi2019visually} and VC-PCFG~\citep{zhao-titov-2020-visually}, two state-of-the-art multimodal grammar induction methods that incorporate visual signals from paired image-caption data. In VG-NSL and VC-PCFG, a visual matching loss between representations of images and their captions serves as a regularizer during grammar induction.

\textbf{Setup.} We follow the experimental setup of \citet{zhao-titov-2020-visually}, evaluating on the same splits of the MSCOCO 2014 dataset~\citep{lin2014microsoft}. (Because MSCOCO does not provide captions for their test set, a portion of the validation set is used as a held-out test set.) Images in the MSCOCO dataset are each associated with 5 captions. The final dataset consists of 82,783 training, 1,000 validation, and 1,000 test images.
During preprocessing, all sentences are converted to lowercase and numbers are replaced with the letter "N". For models using word embedding matrices, the most frequent 10,000 words (based on white-space tokenization) are maintained with all other words mapped to a special UNK token. Captions greater than 45 words in length are removed. For \lmethodname{}, we preprocess the dataset by extracting token-level embeddings for each caption from the last layer of an LLM.

\paragraph{Evaluation.} Because the MSCOCO dataset does not have annotated ground truth parse trees, we follow prior work and use a supervised neural parser, Benepar~\citep{kitaev-klein-2018-constituency}, to generate parse trees for evaluation. Each unsupervised grammar induction method is evaluated by computing the F1 score between the predicted parse tree and the parse tree generated by Benepar. Due to instabilities observed during training, each method is trained with 10 random seeds and then the mean and standard deviation over the top 4 seeds (based on validation F1) are reported.

\paragraph{Implementation.} For baseline models we use the implementation and hyperparameters provided by \citet{zhao-titov-2020-visually}.\footnote{https://github.com/zhaoyanpeng/vpcfg}

The original implementation of VC-PCFG uses a ResNet-152 network to embed images. However, there are now image embedding networks that are stronger than ResNet-152. To provide a fair comparison between our text-only model and multimodal approaches, we improve VC-PCFG by replacing ResNet-152 with ResNetV1.5 - 152~\citep{ResNetV1.5}. We also improve the optimization hyperparameters (learning rate and network dropout). We refer to the modernized version of VC-PCFG as VC-PCFG++. VC-PCFG++ outperforms VC-PCFG by about $3$ points in both corpus and sentence-level F1 scores.   

For \lmethodname{}, we use an OPT-2.7B~\citep{zhang2022opt} model to extract token-level embeddings for each sentence. Sentence embeddings are then mean-pooled and passed through a single linear layer inference network. We use dropout of 0.5 on both the mean-pooled sentence embedding and the output latent vector from the inference network. 

\subsubsection{Results} 
\label{sec:image_results}

Table \ref{tab:eval_mscoco} shows test F1 scores for each model. \lmethodname{} achieves the highest overall corpus-level F1 (C-F1) and sentence-level F1 (S-F1) scores. Note that \lmethodname{} does not use paired visual features, and contains 85\% fewer parameters than the previous state-of-the-art approach (VC-PCFG).

\label{sec:video_results}
 \begin{table*}[t]
\centering
\caption{\textbf{Grammar induction with video and text}. Comparison across three video-text parsing benchmark datasets (DiDeMo, YouCook2 \& MSRVTT). We show performance of simple rule-based baselines (top), prior state-of-the-art multimodal methods (middle) and text-only models including ours (\lmethodname) (bottom). \lmethodname{} outperforms all prior methods.}
    \label{tab:video_results}
\resizebox{1\linewidth}{!}{
\setlength{\tabcolsep}{4.0pt}
\small
    \centering
    \begin{tabular}{lll|cc|ll|ll|ll}
    \toprule
	\phantom{text} & \multicolumn{2}{l}{\multirow{3}*{PCFG Method}} & \multirow{3}*{LLM}  &  \multirow{3}*{RGB} & \multicolumn{2}{c|}{DiDeMo} & \multicolumn{2}{c|}{YouCook2}  & \multicolumn{2}{c}{MSRVTT} \\
	\cmidrule(lr){6-7}
	\cmidrule(lr){8-9}
	\cmidrule(lr){10-11}
	& \multicolumn{2}{c}{} & & &C-F1 & S-F1 & C-F1 & S-F1 & C-F1 & S-F1 \\
    \midrule
    \multicolumn{10}{l}{\textit{Rule-based baselines}} & \\ 
    \rule{0pt}{\normalbaselineskip} 
    & \multicolumn{2}{l}{Left Branching}&No& No & $16.2$ & $18.5$ & $6.8 $ & $5.9 $ & $14.4$ & $16.8$ \\
    & \multicolumn{2}{l}{Right Branching}&No& No &$53.6$ & $57.5$ & $35.0$ & $41.6$ & $54.2$ & $58.6$ \\
    & \multicolumn{2}{l}{Random}&No& No&$29.4_{{\stdev{0.3}}}$ & $32.7_{{\stdev{0.5}}}$ & $21.2_{{\stdev{0.2}}}$ & $24.0_{{\stdev{0.2}}}$ & $27.2_{{\stdev{0.1}}}$ & $30.5_{{\stdev{0.1}}}$ \\
    \midrule 
    \multicolumn{10}{l}{\textit{Methods using extra-linguistic inputs}} & \\ 
    \rule{0pt}{\normalbaselineskip} 
    & \multicolumn{2}{l}{VC-PCFG~\citep{zhao-titov-2020-visually}} & No& Yes & $42.2_{\stdev{12.3}}$ & $43.2_{\stdev{14.2}}$ & $42.3_{\stdev{5.7}}$ & $47.0_{\stdev{5.6}}$ & $49.8_{\stdev{4.1}}$ & $54.2_{\stdev{4.0}}$ \\
    & \multicolumn{2}{l}{MMC-PCFG~\citep{zhang2021video}} &No& Yes & ${55.0}_{\stdev{3.7 }}$ & ${58.9}_{\stdev{3.4 }}$ & ${44.7}_{\stdev{5.2}}$ & ${48.9}_{\stdev{5.7}}$ & $56.0_{\stdev{1.4}}$ & $60.0_{\stdev{1.2}}$ \\
    \midrule
   \multicolumn{10}{l}{\textit{Methods using only textual inputs}} & \\ 
    \rule{0pt}{\normalbaselineskip} 
    & \multicolumn{2}{l}{C-PCFG~\citep{kim-etal-2019-compound}} &No& No & $38.2_{\stdev{5.0}}$ & $40.4_{\stdev{4.1}}$ & $37.8_{\stdev{6.7}}$ & $41.4_{\stdev{6.6}}$ & $50.7_{\stdev{3.2}}$ & $55.0_{\stdev{3.2}}$ \\
    & \multicolumn{2}{l}{\textbf{LC-PCFG} (Ours)} &Yes& No & $\mathbf{57.1}_{\stdev{4.7 }}$ & $\mathbf{60.0}_{\stdev{5.2}}$ & $\mathbf{52.4}_{\stdev{0.1}}$ & $\mathbf{57.7}_{\stdev{0.1}}$ & $\mathbf{56.1}_{\stdev{3.6}}$ & $\mathbf{61.2}_{\stdev{3.7}}$ \\
    \bottomrule
    \end{tabular}
    }

\vspace{-0.1in}
\end{table*}
\subsection{Video-assisted Parsing}
\label{subsec:video-assisted-parsing}

The results in Table \ref{tab:eval_mscoco} show that a text-only approach can outperform approaches that incorporate multimodal inputs from images. However, some have argued that images are a static snapshot of the world, and therefore may lack information needed to induce verb phrases~\citep{zhang2021video}. Based on the intuition that video can provide better multimodal training signals, one study presented an approach for grammar induction (MultiModal Compound PCFG; MMC-PCFG) that incorporates both visual and auditory signals from videos ~\citep{zhang2021video}. MMC-PCFG aggregates multimodal features and achieved a substantial improvement over previous multimodal methods for grammar induction. To test whether a text-only baseline can achieve the same improvements as a video-enhanced method, we compare \lmethodname{} to MMC-PCFG.

\paragraph{Setup.} Following \citet{zhang2021video}, we use three benchmarking video datasets for our experiments: Distinct Describable Moments (\textit{DiDeMo})~\citep{anne2017localizing}, Youtube Cooking (\textit{YouCook2})~\citep{zhou2018towards} and MSRVideo to Text (\textit{MSRVTT})~\citep{xu2016msr}. DiDeMo consists of unedited, personal videos in diverse visual settings with pairs of localized video segments and referring expressions. It includes 32994, 4180 and 4021 video-sentence pairs in the training, validation, and test sets. YouCook2 contains 2000 videos that are nearly equally distributed over 89 recipes. Each video contains 3–16 procedure segments. It includes 8713, 969 and 3310 video-sentence pairs in the training, validation and test sets. MSRVTT is a  large-scale benchmark for video understanding with 10K web video clips with 41.2 hours and 200K clip-sentence pairs in total. It includes 130260, 9940 and 59794 video-sentence pairs across all the data splits.

The extracted multimodal features~\citep{zhang2021video} include object features (SENet~\citep{xie2017aggregated}), action features (I3D~\citep{carreira2017quo}), scenes~\citep{huang2017densely,zhou2017places}, audio~\citep{hershey2017cnn}, OCR~\citep{deng2018pixellink,liu2018synthetically}, faces~\citep{liu2016ssd,he2016identity} and speech~\citep{mikolov2013efficient}. 

We run all experiments 4 times for 10 epochs each, with different random seeds. We report the mean and standard deviation of the C-F1 and S-F1 scores.

\subsubsection{Results} 

Table~\ref{tab:video_results} compares grammar induction performance between C-PCFG, VC-PCFG (which incorporates visual signals), and MMC-PCFG (which incorporates signals from multiple extralinguistic modalities). \lmethodname{} outperforms the video-regularized models for all three benchmark datasets.

\subsection{Large-scale Video Pretraining}
\begin{table*}[t]
\centering
\caption{\textbf{Transferring Learnt Grammar}. Models are trained on the `Trainset' data and evaluated without additional training on the target benchmarks (DiDeMO, YouCook2 \& MSRVTT) on the Sentence-level F1 (S-F1) and Corpus-level F1 (C-F1) metrics. All HowTo100M results are reported on 592k samples.} 
\label{tab:video_result_million}
\resizebox{1\linewidth}{!}{
\small
\centering

\begin{tabular}{lc|cc|cc|cc}
\toprule
\multirow{2}{*}{Method} & \multirow{2}{*}{Trainset}& \multicolumn{2}{c|}{DiDeMo} & \multicolumn{2}{c|}{YouCook2} & \multicolumn{2}{c}{MSRVTT}\\
\cmidrule(lr){3-4}
\cmidrule(lr){5-6}
\cmidrule(lr){7-8}
& & C-F1 & S-F1 & C-F1 & S-F1 & C-F1 & S-F1 \\
\midrule
MMC-PCFG & DiDeMo & ${55.0}_{\stdev{3.7}}$ & ${58.9}_{\stdev{3.4}}$ & $49.1_{\stdev{4.4}}$ & $53.0_{\stdev{4.9}}$ & $49.6_{\stdev{1.4}}$ & $53.8_{\stdev{0.9}}$ \\
MMC-PCFG & YouCook2 & $40.1_{\stdev{4.4}}$ & $44.2_{\stdev{4.4}}$ & ${44.7}_{\stdev{5.2}}$ & ${48.9}_{\stdev{5.7}}$ & $34.0_{\stdev{6.4}}$ & $37.5_{\stdev{6.8}}$ \\ 
MMC-PCFG& MSRVTT & $59.4_{\stdev{2.9}}$ & ${62.7}_{\stdev{3.3}}$ & $49.6_{\stdev{3.9}}$ & $54.2_{\stdev{4.1}}$ & $56.0_{\stdev{1.4}}$ & $60.0_{\stdev{1.2}}$ \\
MMC-PCFG & HowTo100M & $58.5_{\stdev{7.3}}$ & $62.4_{\stdev{7.9}}$ & $53.9_{\stdev{6.6}}$ & $58.0_{\stdev{7.1}}$ & $55.1_{\stdev{7.0}}$ & $60.2_{\stdev{8.0}}$ \\
PTC-PCFG & HowTo100M & $\mathbf{61.3}_{\stdev{3.9}}$ & $\mathbf{65.2}_{\stdev{5.3}}$ & ${58.9}_{\stdev{2.5}}$ & ${63.2}_{\stdev{2.3}}$ & ${57.4}_{\stdev{4.6}}$ & ${62.8}_{\stdev{5.7}}$ \\
\textbf{LC-PCFG} (Ours) & HowTo100M & $60.6_{\stdev{5.2}}$ &$61.5_{\stdev{6.1}}$& $\mathbf{61.1}_{\stdev{2.1}}$ & $\mathbf{65.2}_{\stdev{1.4}}$ & $\mathbf{59.4}_{\stdev{5.0}}$ & $\mathbf{63.0}_{\stdev{5.8}}$  \\
\bottomrule
\end{tabular}
}

\vspace{-0.2in}
\end{table*}
While MMC-PCFG incorporates multimodal inputs from small amounts of video data, other work has proposed to use larger scale video data for improve grammar induction~\citep{zhang2022training}. That work proposed Pre-Trained Compound PCFGs (PTC-PCFG), a multimodal method for grammar induction that obtains paired video and text inputs from captioned instructional YouTube videos in the \textit{HowTo100M} dataset~\citep{miech2019howto100m}. Then a matching loss between these paired inputs is used as a regularizer during grammar induction. PTC-PCFG outperformed previous state-of-the-art multimodal grammar induction models.

To determine how PTC-PCFG compares to our text-only baseline, we train \lmethodname{} with the captions of the \textit{HowTo100M}dataset~\citep{miech2019howto100m} without using any multimodal inputs. Following~\citet{zhang2022training}, we induce a grammar from 592k samples of the HowTo100M train set and then evaluate on the three video-enhanced parsing benchmarks shown in Table~\ref{tab:video_results} (DiDeMo, YouCook2, and MSRVTT).

Table~\ref{tab:video_result_million} shows the test F1 scores for MMC-PCFG, PTC-PCFG, and \lmethodname{} on the three video-enhanced parsing benchmarks. \lmethodname{} outperforms MMC-PCFG, even in settings where \lmethodname{} is trained on out-of-distribution HowTo100M dataset and MMC-PCFG is trained on in-distribution samples from each benchmark dataset. On the three benchmarks, \lmethodname{} either outperforms or nearly matches PTC-PCFG.

\subsection{Runtime Comparison} 
To compare the runtime of each method, we follow the setting of PTC-PCFG and calculate the time to extract embeddings and train each model. Table~\ref{tab:running_cost2} shows the runtime for each model. \lmethodname{} requires more time for embedding extraction than VC-PCFG, but \lmethodname{} results 10$\times$ less time for embedding extraction time compared to video-enhanced models. \lmethodname{} is 1.3 to 8.8 times faster to train than either image-enhanced or video-enhanced models.
\begin{table}[t]
\centering
\centering
\small
\caption{\textbf{Training Time Evaluation} for both image-based (top) and video-based (bottom) grammar induction methods. Run-time for both pre-extracting the embeddings (`Embedding') and model training (`Training') are reported. We pre-embed captions for LC-PCFG with two 24GB Titan RTX GPUs and pre-embed images/videos for models with a visual component. Training times for image and video results are benchmarked on a single 12G 2080 Ti and on 2$\times$ 32G V100s respectively. }
\label{tab:running_cost2}
\begin{tabular}{lcc}
\toprule
     PCFG Method &   Embedding (hours) &  Training (hours) \\
 \midrule
     C-PCFG  & - & $7.6$  \\
     VC-PCFG & $0.25$ & $13.3$  \\
     LC-PCFG & $2.0$ & $8.0$\\
 \midrule 
     C-PCFG  & - & $1.5$  \\
     MMC-PCFG & >$25$ & $15$  \\
     PTC-PCFG & >$25$ & $10$  \\
     LC-PCFG (Ours) & $2.5$ & $1.7$\\
 \bottomrule
\end{tabular}

\end{table}

\section{Model Analysis}
\label{sec:ablation}

\subsection{Perplexity-based Evaluation}

To facilitate comparisons between methods, the results reported in Section \ref{sec:experiments} are based on the model selection procedure used in prior studies~\citep{zhao-titov-2020-visually,zhang2021video,zhang2022training}. This model selection procedure trains models with different random seeds, and then uses validation C-F1 score to choose a subset of random seeds for test evaluation.

However, this model selection procedure assumes that gold parse trees are available during validation. To ensure that our results do not rely on having gold trees at validation time, we repeat our experiments but instead use perplexity (PPL)~\citep{chen1998evaluation} to perform model selection:
$$ PPL(X) = -\frac{1}{t}\sum_{i}^{t}log\ p(x_i|x_{<i})$$
where $X=(x_1, x_2, x_3,...,x_t)$ is a tokenized sequence of words and $p(x_i|x_{<i})$ represents the log-likelihood of the ith token conditioned on the preceding tokens ${x_{<i}}$.

PPL allows us to perform model selection without relying on gold parse trees. We train models with 10 random seeds, and then use PPL to select the four best-performing seeds. 

Table~\ref{tab:image_unsupervised_selection} shows test C-F1 performance on image-assisted parsing for experiments in which we used PPL to perform model selection. \lmethodname{} consistently outperforms methods that use multimodal inputs. We observe the same results for video-assisted parsing (Table~\ref{tab:video_result_ppl}).

\label{sec:perplexity_evaluation}
\begin{table}[h]
\centering

\caption{\textbf{Unsupervised Run Selection Criterion for Unsupervised Grammar Induction}. Corpus-level F1 scores using validation set F1 (`Val-F1'), perplexity (`PPL'), and mean branching factor (`MBF', the average proportion between leaves in the right and left branches of nodes in each tree across the corpus). Unlike Validation-F1 based-selection, PPL and MBF do not require gold trees during validation.}
\small
\begin{tabular}{l|clll}
    \toprule
    \multirow{2}{*}{PCFG  Method}  & \multicolumn{3}{c}{Run Selection Criteria}\\
    \cmidrule(lr){2-5} & \small Val-F1 & \small PPL & MBF \\
     \midrule
     C-PCFG & $60.1_{\stdev{ 4.6}}$ & $52.0_{\stdev{ 7.5}}$ & $56.8_{\stdev{ 9.3}}$ & \\
     VC-PCFG & $61.3_{\stdev{ 2.6}}$ & $55.3_{\stdev{ 10.2}}$ & $51.0_{\stdev{ 13.4}}$ & \\
     \textbf{LC-PCFG} (Ours) & $\mathbf{67.2}_{\stdev{ 1.1}}$ & $\mathbf{67.2}_{\stdev{ 1.1}}$ & $\mathbf{65.3}_{\stdev{ 2.1}}$ & \\
     \bottomrule
\end{tabular}

\label{tab:image_unsupervised_selection}
\vspace{-0.1in}
\end{table}

\begin{table*}[b]
\centering
\caption{
  \textbf{Unsupervised Run Selection Criterion for Unsupervised Grammar Induction}. Similar to Table~\ref{tab:image_unsupervised_selection}, we report the results of run selection based on validation perplexity (PPL) for video benchmarks (Section~\ref{sec:perplexity_evaluation}).}
\resizebox{1\linewidth}{!}{
\centering
\small

    \begin{tabular}{lcccccc}
    \toprule
	\multirow{2}*{PCFG Method} & \multicolumn{2}{c}{DiDeMo} & \multicolumn{2}{c}{YouCook2}  & \multicolumn{2}{c}{MSRVTT} \\
	\cmidrule(lr){2-3}
	\cmidrule(lr){4-5}
	\cmidrule(lr){6-7}
	&  C-F1 & S-F1 & C-F1 & S-F1 & C-F1 & S-F1 \\
    \midrule
    Compound~\cite{kim-etal-2019-compound} & $40.4_{\stdev{10.1}}$ & $42.1_{\stdev{9.1}}$ & $38.6_{\stdev{7.2}}$ & $42.8_{\stdev{7.7 }}$ &$49.2_{\stdev{3.8 }}$& $53.1_{\stdev{4.0 }}$   \\
    Multi-modal~\cite{zhang2021video} & $42.1_{\stdev{12.6 }}$ & $45.7_{\stdev{12.4 }}$ & $38.9_{\stdev{3.6 }}$ & $43.8_{\stdev{3.3 }}$ & $48.1_{\stdev{1.0 }}$ & $52.4_{\stdev{0.9 }}$ \\
    \textbf{LC-PCFG} (Ours) &  $\mathbf{46.3}_{\stdev{6.9 }}$ & $\mathbf{49.9}_{\stdev{7.3 }}$& $\mathbf{46.7}_{\stdev{1.1 }}$ & $\mathbf{52.4}_{\stdev{0.8 }}$ & $\mathbf{50.5}_{\stdev{4.0 }}$ & $\mathbf{55.2}_{\stdev{4.4 }}$\\
    \bottomrule
    \end{tabular}
    }

\label{tab:video_result_ppl}
\end{table*}

\subsection{Branching Factor} 
We performed grammar induction over texts in English, which is a right-branching language. To investigate whether induced grammars capture the right-branching nature of English, we measure the branching factor of predicted parse trees. For each branch in each parse tree we measure the proportion of leaves under the right branch over those of the left branch. This proportion is then averaged across all nodes in the tree to produce an average score $s$. $s$ is referred to as the branching factor of the tree ($s >1.0$ means that the tree is overall right-branching, whereas $s<1.0$ means that the tree is overall left-branching). Formally, for each parse tree $t$ with $|t|$ nodes $n\in t$ we compute the mean over nodes' ratio of leaves in their right and left branches: 

\begin{align*}
    \text{MBF}(t) = \frac{1}{|t|}\sum _{n\in t} \frac{\text{CR}(n)}{\text{CL}(n)}  
\end{align*}
where CR and CL are the respective counts of leaves under the right and left branches of a node. 

\begin{table}
    \caption{\textbf{MBF on image-assisted parsing.}}
    \label{tab:parser_branching}
    \centering
    \small
    \begin{tabular}{c|c}
        \toprule
        PCFG Method & MBF  \\
        \midrule
        C-PCFG & $3.4_{\stdev{0.3}}$ \\
        VC-PCFG & $3.4_{\stdev{0.3}}$  \\
        LC-PCFG & $2.5_{\stdev{0.7}}$  \\
        \bottomrule
    \end{tabular}
\end{table}

Table \ref{tab:parser_branching} shows the mean branching factor (MBF) for each model (computed over 10 seeds). We find that all models predict right-branching trees, and \lmethodname{} has the lowest MBF (i.e., most right-branching trees).

To test whether MBF could be used as a run selection criteria, we used MBF instead of validation C-F1 score to select random seeds. For VC-PCFG and LC-PCFG, using MBF as a seed-selection method performs slightly worse than using PPL or validation C-F1 score as a seed selection method.

\subsection{Model Ablations}

To understand the effect of different model components on grammar induction performance, we perform a series of ablations on parsers trained on the MSCOCO dataset.

To understand the contribution of the latent variable $z$, we ablate $z$ in both training and in evaluation.

First we perform inference-time ablations. During inference-time we zero out the latent variable $z$ (`Zero-$z$'), or randomly shuffle $z$ within an evaluation batch (`Random-$z$'). Next we perform training-time ablations. We train a C-PCFG model without latents (`Zero-Train', a vanilla neural PCFG model). 

The performance of each ablated model is shown in Table \ref{tab:ablations}. We find that inference-time ablations on the latent yield comparable performance to the default parsers, whereas omitting the latent during training yields reduced performance from the standard C-PCFG/LC-PCFG models. 
These results suggest that the latent variable may be largely ignored at inference time, but that it serves an important role in the learning process of the parser.

\begin{table}[b!]
\vspace{-0.2in}
\caption{\textbf{Parser Ablations.} Corpus-level F1 scores for PCFG parsers under ablations. We compare the default formulations (`Default') to conditions zeroing out the latent $z$ (`Zero-$z$'), randomly shuffling latents across a batch (`Random-$z$'), shuffling words in each caption (`Shuffle') or zeroing captions out (`Zero-C'), as well as zeroing out latents during training (`Zero-Train'). Note that C-PCFG and LC-PCFG are functionally equivalent in the Zero-Train condition because LLM features are only used in latent computation.} 
    \label{tab:ablations}
    \centering
    \small
    \begin{tabular}{l|ccc}
        \toprule
        \multirow{2}*{Ablation} & \multicolumn{3}{c}{Test Corpus F1} \\
        \cmidrule(lr){2-4}        
        & C-PCFG & VC-PCFG & LC-PCFG \\
        \midrule
        Default & $60.1_{\stdev{ 4.6}}$ & $61.3_{\stdev{ 2.6}}$ & $67.2_{\stdev{ 1.1}}$ \\
        Zero-$z$ & $60.3_{\stdev{ 5.2}}$ & $60.6_{\stdev{ 2.6}}$ & $67.2_{\stdev{ 1.1}}$ \\
        Random-$z$ & $60.3_{\stdev{ 5.2}}$ & $60.9_{\stdev{ 2.5}}$ & $67.2_{\stdev{ 1.1}}$ \\
        Shuffle & $30.0_{\stdev{ 0.7}}$ & $31.0_{\stdev{ 0.9}}$ & $40.6_{\stdev{ 1.0}}$ \\
        Zero-C & $35.2_{\stdev{ 16.1}}$ & $44.6_{\stdev{ 7.6}}$ & $48.6_{\stdev{ 7.5}}$ \\
        Zero-Train & $57.1_{\stdev{6.5}}$ & $58.8_{\stdev{0.9}}$ & $57.1_{\stdev{6.5}}$ \\
        \bottomrule
    \end{tabular}
\end{table}

Lastly, we ablate the input sentences. We evaluate parsers when shuffling (`Shuffle') or zeroing out input caption embeddings (`Zero-C'), word embeddings for VC-PCFG or LLM embeddings for LC-PCFG). We find that ablating the input sentences substantially reduces test performance, suggesting that learned parsers do not merely degenerate to a learned prior at inference time. 

\subsection{Re-adding Visual Signals to \lmethodname{}}
Section \ref{sec:experiments} showed that \lmethodname{} outperforms previous multimodal approaches to grammar induction. But can re-adding visual signals to \lmethodname{} further improve grammar induction? Such an improvement would suggest that multimodal signals contribute to grammar induction beyond what can be learned from text alone.

To test this possibility we re-trained \lmethodname{} with the addition of paired visual features. Visual features were incorporated with the same multimodal regularization loss as used in prior work \cite{shi2019visually,zhao-titov-2020-visually}. Table \ref{tab:llm_plus_image} and Table \ref{tab:llm_plus_video} show the effect of adding image and video signals to \lmethodname{}.

\begin{table}
\centering
\small
\begin{tabular}{cccc}
\toprule
Method  & Params (M) & C-F1 & S-F1 \\
\midrule
 Ours & $\textbf{6.2}$ & $\textbf{67.2}_{\stdev{1.1}}$ & $\textbf{67.8}_{\stdev{1.2}}$ \\
 Ours + ImgFeas  & 32.3 & $59.2_{\stdev{0.5}}$ & $59.4_{\stdev{0.5}}$\\
\bottomrule
\end{tabular}
\caption{
    \label{tab:llm_plus_image}
    \textbf{Adding visual features to \lmethodname{}.} Incorporating visual features (``ImgFeas") into \lmethodname{} degrades performance. 
}
\end{table}

\begin{table*}[h]
\centering
\small
\begin{tabular}{c|cc|cc|cc}
\toprule
 & \multicolumn{2}{c|}{DiDeMo} & \multicolumn{2}{c|}{YouCook2}  & \multicolumn{2}{c}{MSRVTT} \\
 \cmidrule(lr){2-3}
\cmidrule(lr){4-5}
\cmidrule(lr){6-7}
PCFG Method &C-F1 & S-F1 & C-F1 & S-F1 & C-F1 & S-F1 \\
\midrule
Ours & ${57.1}_{\stdev{4.7 }}$ & ${60.0}_{\stdev{5.2}}$ & ${52.4}_{\stdev{0.1}}$ & ${57.7}_{\stdev{0.1}}$ & ${56.1}_{\stdev{3.6}}$ & ${61.2}_{\stdev{3.7}}$ \\
Ours +  VideoFeas & ${50.1}_{\stdev{3.7 }}$ & ${52.9}_{\stdev{3.7}}$ & ${53.2}_{\stdev{1.1}}$ & ${58.0}_{\stdev{0.8}}$ & $51.5_{\stdev{0.5}}$ & $55.9_{\stdev{1.2}}$ \\
\bottomrule
    \end{tabular}
\caption{
    \label{tab:llm_plus_video}
    Incorporating video features (``VideoFeas") into \lmethodname{} degrades performance. 
}
\end{table*}

Adding visual signals to \lmethodname{} reduces performance compared to the text-only version of the model. We observe this degradation across all datasets, both for pixel-based and video-based visual features. We hypothesize that \lmethodname{} may overfit to the added visual features and thereby obfuscate the signals in LLM embeddings.

\section{Conclusion and Future Work}
We propose \lmethodname{}, a strong text-only baseline for grammar induction. \lmethodname{} is a C-PCFG model that incorporates representations from LLMs trained on text alone. On four benchmarks for multimodal grammar induction, \lmethodname{} outperforms several prior state-of-the-art multimodal approaches. Furthermore, adding visual inputs to \lmethodname{} does not improve grammar induction. These experiments show that for grammar induction, the benefits of multimodal inputs can be achieved by more textual data. Our results challenge the notion that multimodal inputs are necessary for grammar induction.

Based on the result that \lmethodname{} performs as well as methods trained on multimodal inputs, we speculate that representations from LLMs provide information that is redundant with information provided by multimodal inputs. Indeed, some work has shown that multimodal inputs improve grammar induction by providing signals of noun concreteness \cite{Kojima2020WhatIL}. Other work has shown that LLMs acquire some knowledge of word concreteness \citep{ramakrishnan-deniz-2021-non}. Thus, large amounts of textual training data may provide signals of word concreteness that obviate multimodal inputs for grammar induction.

\section{Broader Impacts Statement}
Our experiments show that a text-only baseline can outperform computationally intensive multimodal approaches for grammar induction. These results emphasize the promise of less computationally demanding methods, and we we hope they encourage the community to re-think the necessity of expensive multimodal approaches for certain tasks.

\section{Limitations}
Our results show that a strong LLM-based text-only baseline outperforms current state-of-the-art multi-modal grammar induction methods, and that adding visual features to this baseline does not further improve grammar induction. It is possible that future work will find better methods of combining visual features with LLMs, and that these methods will outperform any text-only approaches.

\section{Acknowledgements}
We would like to thank the anonymous reviewers and metareviewers for their constructive comments and suggestions. We also thank the Machine Common Sense project, ONR MURI award number N00014-21-1-2801, and Google's TPU Research Cloud (TRC) for providing cloud
TPUs. RC was funded by the DARPA SemaFor program, BAIR Commons, and an NSF Graduate Research Fellowship. CC was funded by an IBM PhD fellowship.


\clearpage
\bibliography{reference}

\clearpage
\appendix
\appendixpage
\label{sec:appendix}

\section{Parser Branching}\label{sec:parser_branching}
As described in Section \ref{sec:perplexity_evaluation}, we compute a measure, which we refer to as the mean branching factor (MBF), of parsers' predisposition towards generating right vs. left branching trees. 
Specifically, for each parse tree $t$ with $|t|$ nodes $n\in t$ we compute the mean over nodes' ratio of leaves in their right and left branches: 
\begin{align*}
    \text{MBF}(t) = \frac{1}{|t|}\sum _{n\in t} \frac{\text{CR}(n)}{\text{CL}(n)}  
\end{align*}
where CR and CL are the respective counts of leaves under the right and left branches of a node. 

In Table \ref{tab:parser_branching} we present the average MBF (across 10 seeds) for each PCFG method. 
Additionally, we compute the Pearson correlation (PC) between test-set corpus-level F1 performance and parser MBF. 
As may be observed, there is generally a negative correlation between CF1 and MBF, with the text-only C-PCFG and LC-PCFG methods displaying greater magnitudes of negative correlation. 
These results suggest that the MBF may also be used as an unsupervised run-selection metric during validation. 

\begin{figure}[!h]
    \centering
    \includegraphics[width=.9\linewidth]{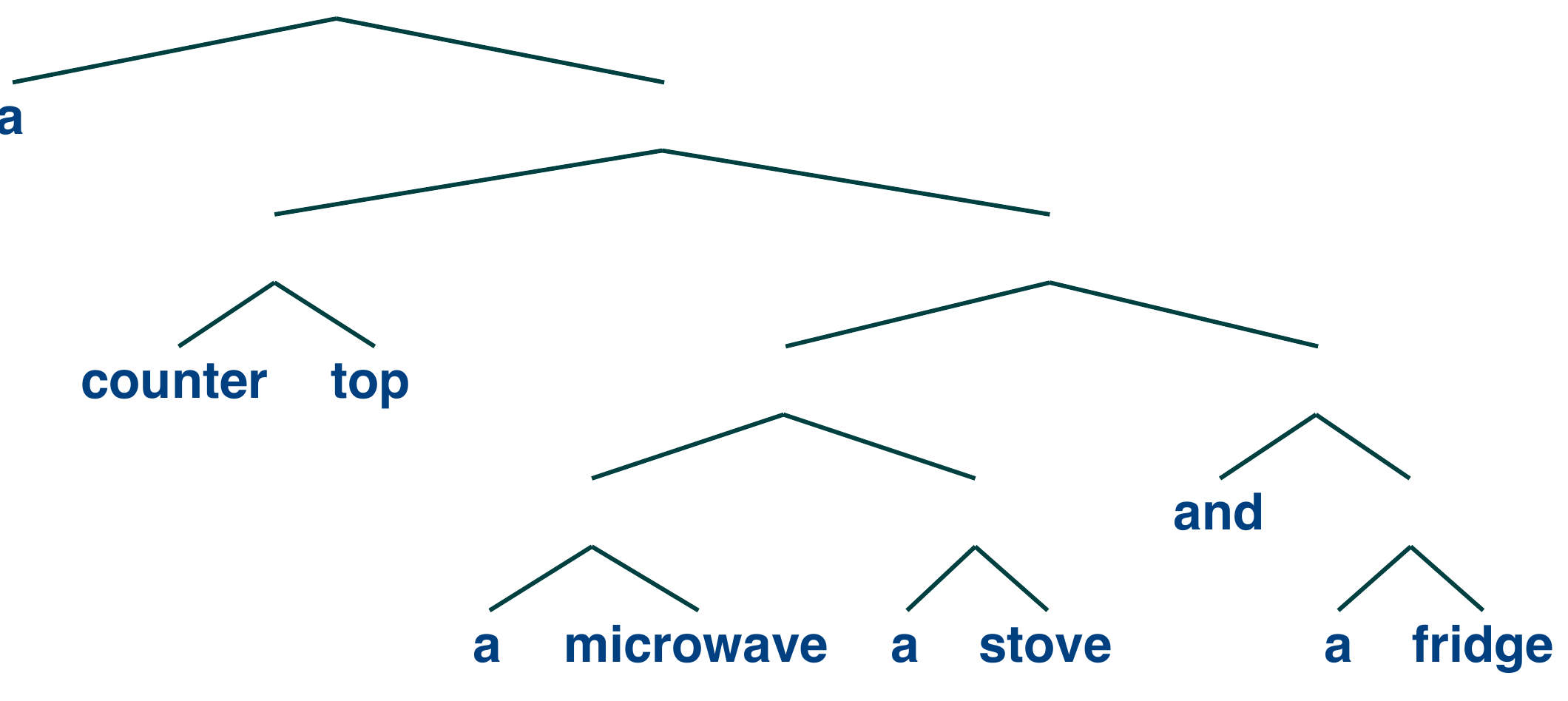}
    \includegraphics[width=.9\linewidth]{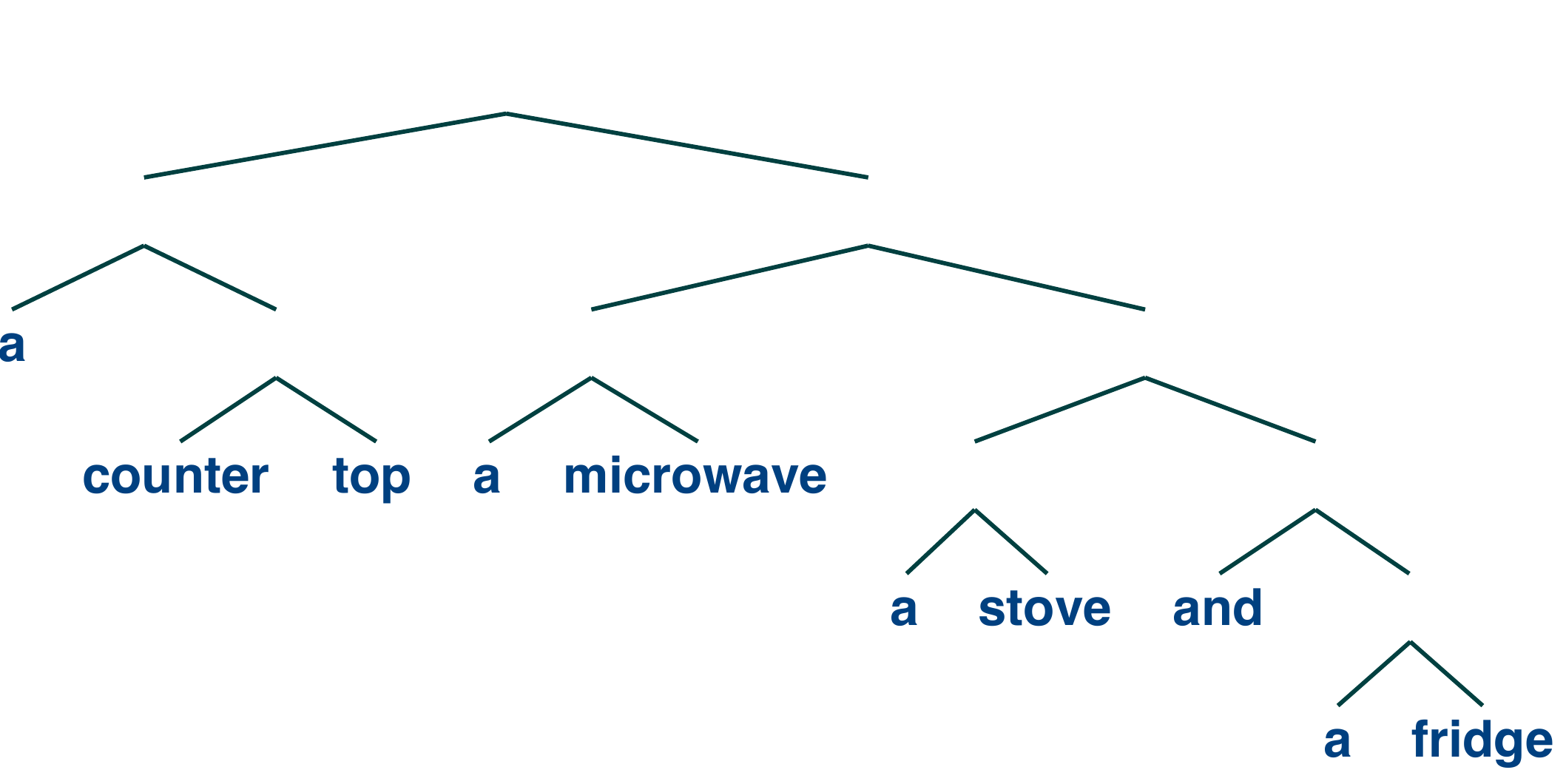}
    \caption{\textbf{Top}: A parse tree generated by \lmethodname{}. \textbf{Bottom}: The gold-parse tree. Note how the parser is generally able to form noun phrases. We note that all parsers in our experiments learn a right-branching bias, which follows intuition since English is a right-branching language.}
    \label{fig:parse_tree}
\end{figure}

In Figure \ref{fig:parse_tree} we show a qualitative example of a parse tree predicted by LC-PCFG and a binarized gold tree generated by Benepar~\cite{kitaev-klein-2018-constituency}. 
We find that all parsers we train produce right-branching grammars.

\begin{table}[!h]
    \centering
    \small
    \begin{tabular}{c|cc}
        \toprule
        PCFG Method & MBF & PC \\
        \midrule
        C-PCFG & $3.4_{\stdev{0.3}}$ & $-0.62$\\
        VC-PCFG & $3.4_{\stdev{0.3}}$ & $-0.12$ \\
        LC-PCFG & $2.5_{\stdev{0.7}}$ & $-0.75$ \\
        \bottomrule
    \end{tabular}
    \caption{\textbf{Parser Branching.} We show the mean branching factor  (\textbf{MBF}) and Pearson correlation (\textbf{PC}) between MBF and corpus-level F1 score for each model. We find that there is a negative correlation between MBF and F1 scores, suggesting that controlling for more balanced trees may be a viable unsupervised run selection strategy for LC-PCFG parsers.}
    \label{tab:parser_branching}
\end{table}

\begin{table*}[!t]
\centering
\caption{Comparison with multi-modalities on three benchmark datasets. Note that our method, using features from OPT-2.7B, yields superior results despite not being regularized by multiple modalities. "W/ MD" denotes using multimodal data.}
\label{tab:video_results_full}

\resizebox{1.04\linewidth}{!}{
\setlength{\tabcolsep}{4.0pt}
\small
    \centering
    \begin{tabular}{cccccccccccc}
    \toprule
	\multicolumn{2}{c}{\multirow{2}*{Method}}&\multirow{2}{*}{W/ MD} & \multicolumn{5}{c}{DiDeMo} & \multicolumn{2}{c}{YouCook2}  & \multicolumn{2}{c}{MSRVTT} \\
	\cmidrule(lr){4-8}
	\cmidrule(lr){9-10}
	\cmidrule(lr){11-12}
	\multicolumn{2}{c}{} && NP & VP & PP & C-F1 & S-F1 & C-F1 & S-F1 & C-F1 & S-F1 \\
    \hline
    \multicolumn{2}{c}{LBranch}&\xmark& $41.7$ & $0.1 $ & $0.1 $ & $16.2$ & $18.5$ & $6.8 $ & $5.9 $ & $14.4$ & $16.8$ \\
    \multicolumn{2}{c}{RBranch}&\xmark& $32.8$ & $\mathbf{91.5}$ & ${66.5}$ & $53.6$ & $57.5$ & $35.0$ & $41.6$ & $54.2$ & $58.6$ \\
    \multicolumn{2}{c}{Random} &\xmark& $36.5_{\pm0.6 }$ & $30.5_{\pm0.5 }$ & $30.1_{\pm0.5 }$ & $29.4_{\pm0.3 }$ & $32.7_{\pm0.5 }$ & $21.2_{\pm0.2}$ & $24.0_{\pm0.2}$ & $27.2_{\pm0.1}$ & $30.5_{\pm0.1}$ \\
    \multicolumn{2}{c}{C-PCFG} &\xmark& $\mathbf{72.9}_{\pm5.5 }$ & $16.5_{\pm6.2 }$ & $23.4_{\pm16.9}$ & $38.2_{\pm5.0 }$ & $40.4_{\pm4.1 }$ & $37.8_{\pm6.7}$ & $41.4_{\pm6.6}$ & $50.7_{\pm3.2}$ & $55.0_{\pm3.2}$ \\
    \hline
    \multirow{11}{*}{\rotatebox{90}{VC-PCFG}} 
    & Object &\checkmark& $70.5_{\pm15.3}$ & $25.7_{\pm15.9}$ & $36.5_{\pm24.6}$ & $42.6_{\pm10.4}$ & $44.0_{\pm10.4}$ & $39.9_{\pm8.7}$ & $44.9_{\pm8.3}$ & $52.2_{\pm1.2}$ & $56.0_{\pm1.6}$ \\
    & Action &\checkmark& $57.9_{\pm13.5}$ & $45.7_{\pm14.1}$ & $45.8_{\pm17.2}$ & $45.1_{\pm6.0 }$ & $49.2_{\pm6.0 }$ & $40.6_{\pm3.6}$ & $45.7_{\pm3.2}$ & $54.5_{\pm1.6}$ & ${59.1}_{\pm1.7}$ \\
    & R2P1D &\checkmark& $61.2_{\pm8.5 }$ & $38.1_{\pm5.4 }$ & $62.1_{\pm4.1 }$ & $48.1_{\pm4.4 }$ & $50.7_{\pm4.2 }$ & $39.4_{\pm8.1}$ & $44.4_{\pm8.3}$ & $54.0_{\pm2.5}$ & $58.0_{\pm2.3}$ \\
    & S3DG &\checkmark& $61.3_{\pm13.4}$ & $31.7_{\pm16.7}$ & $51.8_{\pm8.0 }$ & $44.0_{\pm2.7 }$ & $46.5_{\pm5.1 }$ & $39.3_{\pm6.5}$ & $44.1_{\pm6.6}$ & $50.7_{\pm3.2}$ & $54.7_{\pm2.9}$ \\
    & Scene &\checkmark& $62.2_{\pm9.6 }$ & $30.6_{\pm12.3}$ & $41.1_{\pm24.8}$ & $41.7_{\pm6.5 }$ & $44.9_{\pm7.4 }$ & $-$ & $-$ & ${54.6}_{\pm1.5}$ & $58.4_{\pm1.3}$ \\
    & Audio &\checkmark& $64.2_{\pm18.6}$ & $21.3_{\pm26.5}$ & $34.7_{\pm11.0}$ & $38.7_{\pm3.7 }$ & $39.5_{\pm5.2 }$ & $39.2_{\pm4.7}$ & $43.3_{\pm4.9}$ & $52.8_{\pm1.3}$ & $56.7_{\pm1.4}$ \\
    & OCR &\checkmark& $64.4_{\pm15.0}$ & $27.4_{\pm19.5}$ & $42.8_{\pm31.2}$ & $41.9_{\pm16.9}$ & $44.6_{\pm17.5}$ & $38.6_{\pm5.5}$ & $43.2_{\pm5.6}$ & $51.0_{\pm3.0}$ & $55.5_{\pm3.0}$ \\
    & Face &\checkmark& $60.8_{\pm16.0}$ & $31.5_{\pm17.0}$ & $52.8_{\pm9.8 }$ & $43.9_{\pm4.5 }$ & $46.3_{\pm5.5 }$ & $-$ & $-$ & $50.5_{\pm2.6}$ & $54.5_{\pm2.6}$ \\
    & Speech &\checkmark& $61.8_{\pm12.8}$ & $26.6_{\pm17.6}$ & $43.8_{\pm34.5}$ & $40.9_{\pm16.0}$ & $43.1_{\pm16.1}$ & $-$ & $-$ & $51.7_{\pm2.6}$ & $56.2_{\pm2.5}$ \\
    & Concat &\checkmark& $68.6_{\pm8.6 }$ & $24.9_{\pm19.9}$ & $39.7_{\pm19.5}$ & $42.2_{\pm12.3}$ & $43.2_{\pm14.2}$ & $42.3_{\pm5.7}$ & $47.0_{\pm5.6}$ & $49.8_{\pm4.1}$ & $54.2_{\pm4.0}$ \\
    \midrule
    \multicolumn{2}{c}{MMC-PCFG} &\checkmark& $67.9_{\pm9.8 }$ & ${52.3}_{\pm9.0 }$ & $63.5_{\pm8.6 }$ & ${55.0}_{\pm3.7 }$ & ${58.9}_{\pm3.4 }$ & ${44.7}_{\pm5.2}$ & ${48.9}_{\pm5.7}$ & $56.0_{\pm1.4}$ & $60.0_{\pm1.2}$ \\
    \midrule
    \multicolumn{2}{c}{\textbf{LC-PCFG}} &\xmark& ${71.1}_{\pm6.6}$ & $47.4_{\pm12.6}$ & $\mathbf{76.9}_{\pm7.3}$ & $\mathbf{57.1}_{\pm4.7 }$ & $\mathbf{60.0}_{\pm5.2 }$ & $\mathbf{52.4}_{\pm0.1}$ & $\mathbf{57.7}_{\pm0.1}$ & $\mathbf{56.1}_{\pm3.6}$ & $\mathbf{61.2}_{\pm3.7}$ \\
    \bottomrule
    \end{tabular}
    }

\end{table*}

\section{Video-assisted Parsing}
In Table~\ref{tab:video_results_full} we present the unabridged version of comparisons for grammar induction with video and text, corresponding to Table 2 in the paper.

\end{document}